# The Utility of General Domain Transfer Learning for Medical Language Tasks

Authors: Daniel Ranti, B.S.[1], Katie Hanss, A.B.[1], Shan Zhao, M.D., Ph.D.[2], Varun Arvind, B.S.[1], Joseph Titano, M.D.[3], Anthony Costa, Ph.D.[1], Eric Oermann, M.D.*[1]

[1]Department of Neurological Surgery, Mount Sinai Health System, 1 Gustave Levy Pl, New York, NY 10029

[2]Department of Anesthesiology, Mount Sinai Health System, 1 Gustave Levy Pl, New York, NY 10029

[3]Department of Radiology, Mount Sinai Health System, 1 Gustave Levy Pl, New York, NY 10029

## ABSTRACT

The purpose of this study is to analyze the efficacy of transfer learning techniques and transformer-based models as applied to medical natural language processing (NLP) tasks, specifically radiological text classification. We used 1,977 labeled head CT reports, from a corpus of 96,303 total reports, to evaluate the efficacy of pretraining using general domain corpora and a combined general and medical domain corpus with a bidirectional representations from transformers (BERT) model for the purpose of radiological text classification. Model performance was benchmarked to a logistic regression using bag-of-words vectorization and a long short-term memory (LSTM) multi-label multi-class classification model, and compared to the published literature in medical text classification. The BERT models using either set of pretrained checkpoints outperformed the logistic regression model, achieving sample-weighted average F1-scores of 0.87 and 0.87 for the general domain model and the combined general and biomedical-domain model. General text transfer learning may be a viable technique to generate state-of-the-art results within medical NLP tasks on radiological corpora, outperforming other deep models such as LSTMs. The efficacy of pretraining and transformer-based models could serve to facilitate the creation of groundbreaking NLP models in the uniquely challenging data environment of medical text.

## 1 Introduction

Electronic health records (EHR) have become an increasingly important but inaccessible source of medical data, with unstructured notes comprising up to 50% of the accumulated data contained within a given patient's chart[1]. The current gold-standard of data abstraction from physician progress reports, nursing care notes, triage notes, etc. is a time consuming, labor-intensive process that limits the potential usefulness of this rich source of data. The use of administrative data, such as ICD-9 codes, to identify outcomes has been attempted to varying degrees, and is inadequate and inaccurate, requiring information that is often documented only in physician and nursing notes[2–4]. The automated extraction of insights from medical text has long been a source of both commercial and academic research efforts, and has proven its potential in a variety of medical contexts, from analyzing radiological reports to fully classifying and diagnosing pediatric diseases[5–9].

Automated classification and segmentation of textual EHR data have numerous applications in research and clinical tasks that are otherwise untenable by hand. Outcome measures are often poorly codified in EHRs, and identifying the presence or absence of a label generally requires the context of the surrounding note, which includes a diverse set of phrases and language potentially not related to the primary label[4]. Manual assembly of such data is error-prone and time intensive, and for research that necessitates thousands of labeled reports to identify rare events, such as adverse drug reactions, curation by hand is impossible. Efforts by clinical informatics researchers to create software that automatically identifies clinically important events within medical text has been motivated by the prospect of highly accurate, gold-standard labels, and have largely relied on either simple rule-based non-machine learning methods, or more classic natural language processing techniques, such as n-gram models[2,9–11].

Unsupervised pretraining tasks in natural language processing (NLP) are a form of transfer learning that seek to build a broad semantic understanding in order to facilitate improved performance on training and testing tasks, such as text classification[12]. Transfer learning and unsupervised pretraining have proved highly efficacious in benchmark NLP tasks, and hold particular promise in application to medical NLP tasks. Large, publicly available corpora of medical data are limited, in part due to the challenges of properly deidentifying radiological and clinical text. For instance, researchers have found that based on lab values alone, algorithms can predict a patient's identity in future admissions at least 25% of the time[13]. Given this, the ability to leverage nonspecific pretraining tasks to generalize to medical subdomain tasks would be useful in optimizing the limited medical text available. However, commonly used and publicly available datasets, such as the Wikipedia corpus, are not specific to medical subdomain, and it is possible they would not generalize well. Thus, understanding if transfer learning and language models using nonspecific pretraining tasks generalize to medical subdomain tasks is important to the development of future models.

To our knowledge, no one has evaluated the suitability of nonspecific pretraining for medical NLP tasks[14]. While Peng et al. recently proposed the Biomedical Language Understanding Evaluation (BLUE) benchmark and published baseline results for several transfer learning models, all pretraining was performed on medical-domain text (e.g. BioBERT, BERT pretrained on PubMed, BERT pretrained on PubMed and MIMIC-III).[14] In this study, we further the existing body of knowledge by evaluating the capabilities of transfer learning and language models using nonspecific pretraining tasks to generalize to medical subdomain tasks.

## 2 Materials and Methods

Head computed tomographic (CT) scans between 2010 and 2016 were assembled from cases stored within the hospital picture archiving and communication system. In total, 96,303 head CT reports were available, of which 1,977 reports written by neuroradiologists were randomly sampled. Each report was labeled for clinical entities, as per the United Medical Language System Concept Unique Identifier and ordered into a taxonomy (see Figure 1). Three physicians, including two PGY4 radiology residents and a PGY4 neurosurgical resident, generated independent binary labels for each report. Interrater agreement was measured by the Fleiss κ, and, to exclude reports with low agreement, only reports with $\kappa \geq 0.60$ were considered. IRB approval was achieved, and patient consent to use reports was waived. 1,004 of the 1,977 reports have been previously reported[9]. The prior study used a lasso logistic regression to predict labels occurring more than 20 times, whereas this study evaluates the effectiveness of deep learning models and transfer learning on various corpora to predict radiological labels[9].

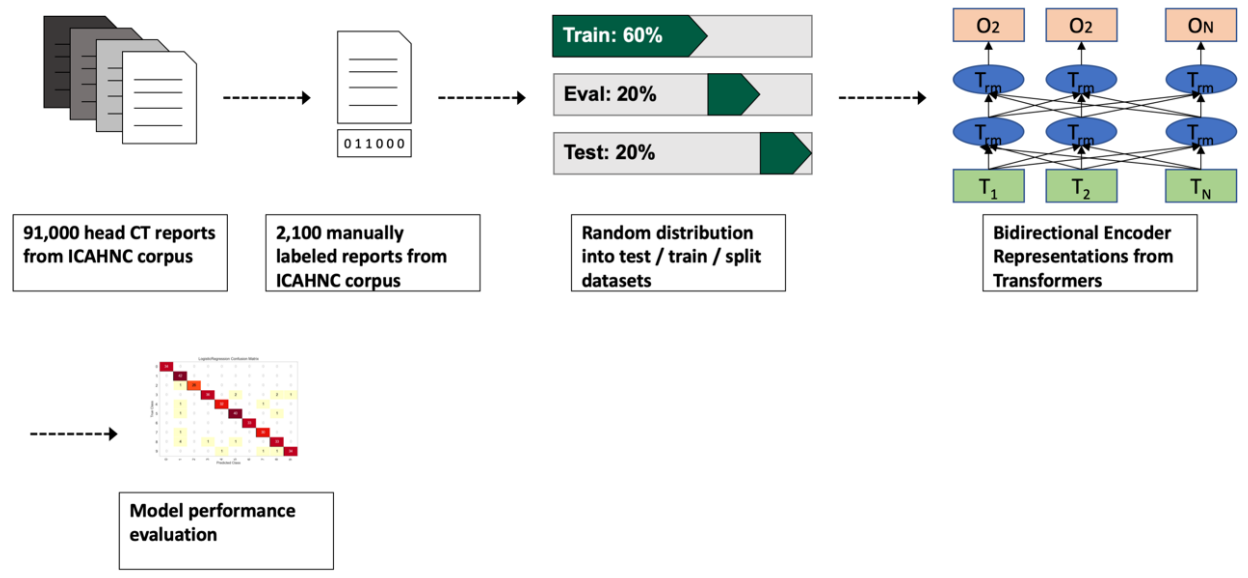

**Figure 1. Flow Diagram:** The overall workflow of the NLP analysis beginning with (A) obtaining raw, unlabeled head CT reports, (B) manual labeling by expert reviewers, (C) pseudo-random, reproducible distribution into training, evaluation, and test datasets (D) training of the model, and (E) model evaluation. ICAHNC = Mount Sinai Hospital and Mount Sinai Queens.

Among the labels, there was significant class imbalance, with the majority of the labels containing only one or two positive instances within the entire labeled corpus (see Figure 2). Of the initial 273, 13 labels with sufficient positive instances were included to evaluate the performance of the model. The cutoff criteria for label inclusion was defined as a label being present in greater than 250 reports. The labels used for the prediction task were as follows: normal, hemorrhage, stroke/infarction/ischemia (infarction, venous thrombosis), vascular abnormality (aneurysm, arteriovenous malformation), chronic small vessel disease, periventricular white matter changes, ventricular abnormalities (i.e. hydrocephalus), atrophy (brain), bone abnormality, bony sinus disease, foreign objects, maxillary sinus disease, and carotid siphon calcification. The reports were pseudo-randomly split into three sets using a pigeonhole function for reproducibility: a training set (59%, 1,161 / 1,977 total reports), an evaluation set (20%, 403 / 1,977 total reports), and a test set (21%, 413 / 1,977 reports).

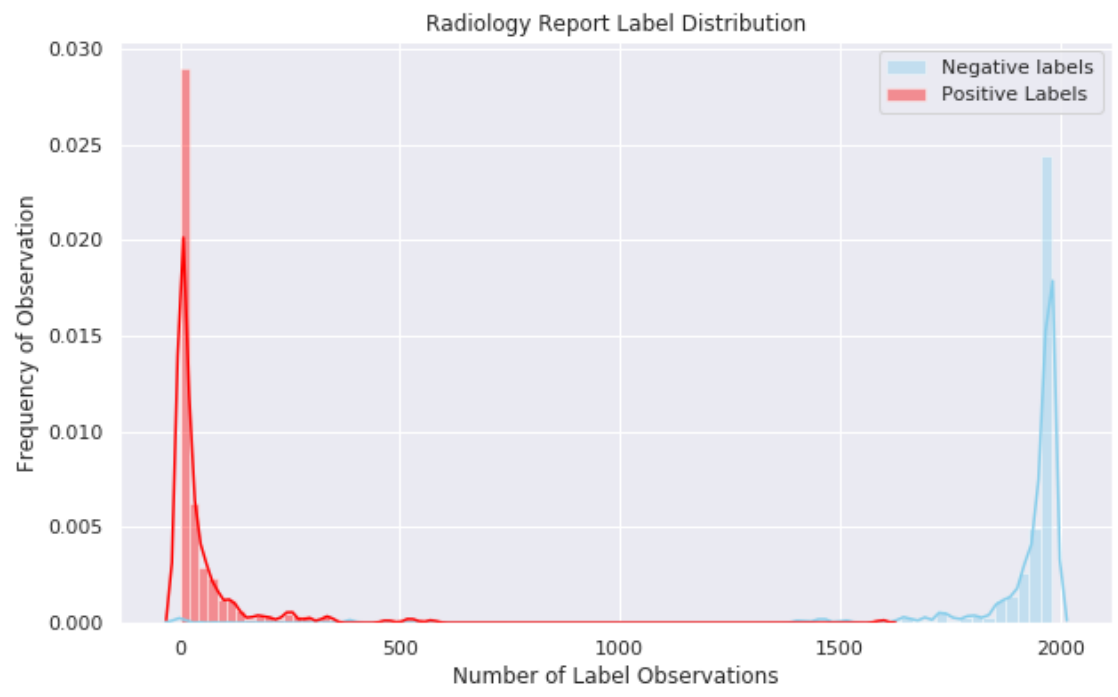

**Figure 2. Label Histogram:** Histogram distribution of positive and negative instances on a per-label basis, showing class imbalance within labels of the original dataset, with the overwhelming majority of labels being negative. The final dataset was chosen by selecting labels that had more than 250 positive instances per report**.**

## 2.1 Natural Language Processing, Text Preprocessing, Transformer Frameworks

NLP is an application of machine learning and deep learning that is specifically geared towards machine interpretations of textual information. The field of NLP has seen rapid improvements in the state-of-the-art on benchmark tests via the use of transfer learning, semi-supervised training and multi-head attention mechanisms[13]. Semi-supervised training refers to a two-stage training procedure, in which the model first learns broad text representations on an unlabeled dataset, and then refines those representations on task-specific labeled data, such as head CT reports[15]. Semi-supervised training is a form of transfer learning, a method in which a model developed on one task is then re-used as a starting point for a second task. Attention refers to a deep learning mechanism that relates varying elements in a textual sequence to best represent the meaning of the sequence as a whole[16].

The model architecture used for this analysis was the base version of the BERT model. The model structure consists of 12 transformer blocks, 768 hidden layers, 12 attention heads, and 110 million total parameters[14]. BERT is unique in that it is based on the transformer architecture, a sequence model that exclusively utilizes soft attention rather than recurrent layers (see Figure 3). Importantly, attention algorithms take into account textual semantics when learning the meaning of a sequence, an advantage not offered by simpler machine learning mechanisms, such as bag-of-words[16]. The BERT model was modified with a 13-dimensional output, one per class, and trained with binary cross entropy with logits in order to obtain an independent probability for each class (multi-label training).

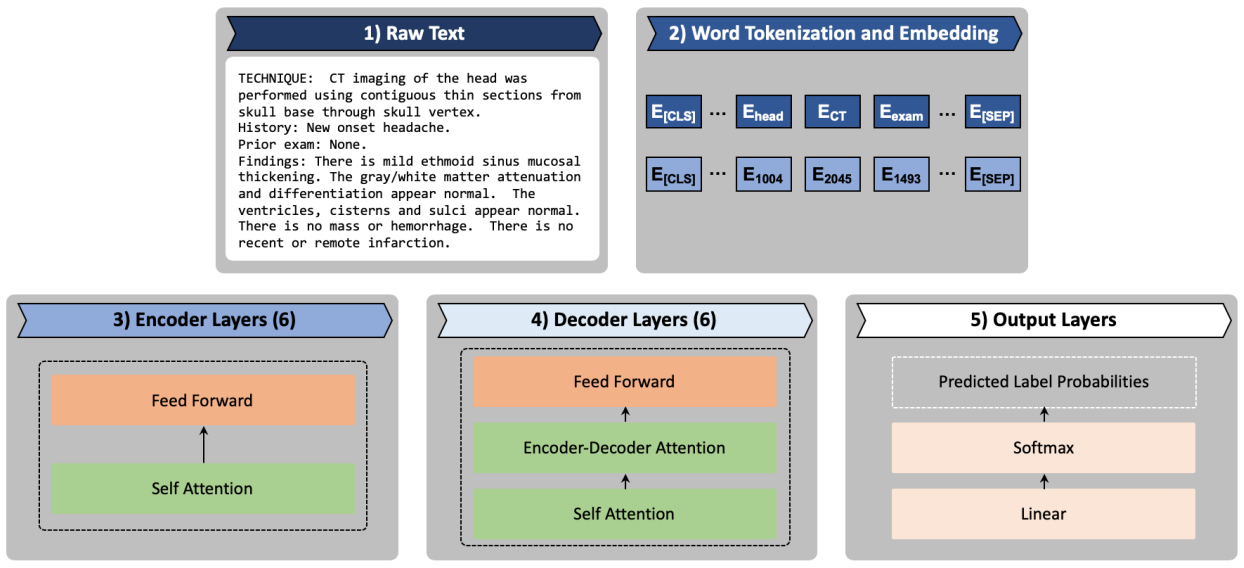

**Figure 3. BERT Diagram:** High level flow of the BERT model beginning with 1) raw text of head CT reports, 2) transfer to lowercase and word tokenization, 3) encoder layers, 4) decoder layers, 5) and finally linear and softmax layers to produce linear probabilities.

The BERT model was chosen due to proven performance on sequence modeling and transduction problems, surpassing other popular mechanisms such as long short-term memory and gated recurrent neural networks, on benchmark NLP tasks. The power of the BERT architecture comes from the use of bidirectional pre-training, in which a cloze deletion task incorporates tokens both to the left and the right of a masked word to be predicted, and a unified architecture across a variety of NLP tasks. By performing the cloze deletion task in an unsupervised manner over a given corpus, the model can build a powerful semantic understanding to then fine-tune for a given task[14]. The utility of a bidirectional approach on a substantial corpus of representative text is evident in the use of a unified architecture to achieve new state-of-the-art results on a diverse set of NLP tasks, including benchmark question answering tasks (Stanford Question Answering Dataset challenge), language comprehension tasks (General Language Understanding Evaluation), and others[14].

Pretraining of the model parameters took place on either the general domain Wikipedia corpus or the combined Wikipedia and PubMed / PMC corpus. Pretraining was accomplished using two unsupervised learning techniques, namely a cloze word prediction task and a next sentence prediction task, on a variety of corpora as previously described[14]. Two separate sets of pretrained parameters were used for the radiological labelling task. The first set of parameters, referred to as general BERT, was based on the unsupervised learning of a combination of two general domain corpora: BooksCorpus (0.8 x $10^9$ words) and English Wikipedia (2.5 x $10^9$ words)[14]. The second set of parameters, referred to as BioBERT, used general BERT as a starting point for further pretraining on two biomedical domain corpora: PubMed abstracts (4.5 x $10^9$ words) and PMC full-text articles (13.5 x $10^9$ words)[17]. Lastly, a random set of pretraining parameters was generated based on the normal distribution, referred to in this article as randomized BERT. These parameters are available to the public and were downloaded from their respective repositories online[13,17].

## 2.2 Statistical Analysis

NLP is an application of machine learning and deep learning that is specifically geared towards machine interpretations of textual information. The field of NLP has seen rapid improvements in the state-of-the-art on benchmark tests via the use of transfer learning, semi-supervised training and multi-head attention mechanisms[13]. Semi-supervised training refers to a two-stage training procedure, in which the model first learns broad text representations on an unlabeled dataset, and then refines those representations on task-specific labeled data, such as head CT reports[15]. Semi-supervised training is a form of transfer learning, a method in which a model developed on one task is then re-used as a starting point for a second task. Attention refers to a deep learning mechanism that relates varying elements in a textual sequence to best represent the meaning of the sequence as a whole (see Figure 4) [16].

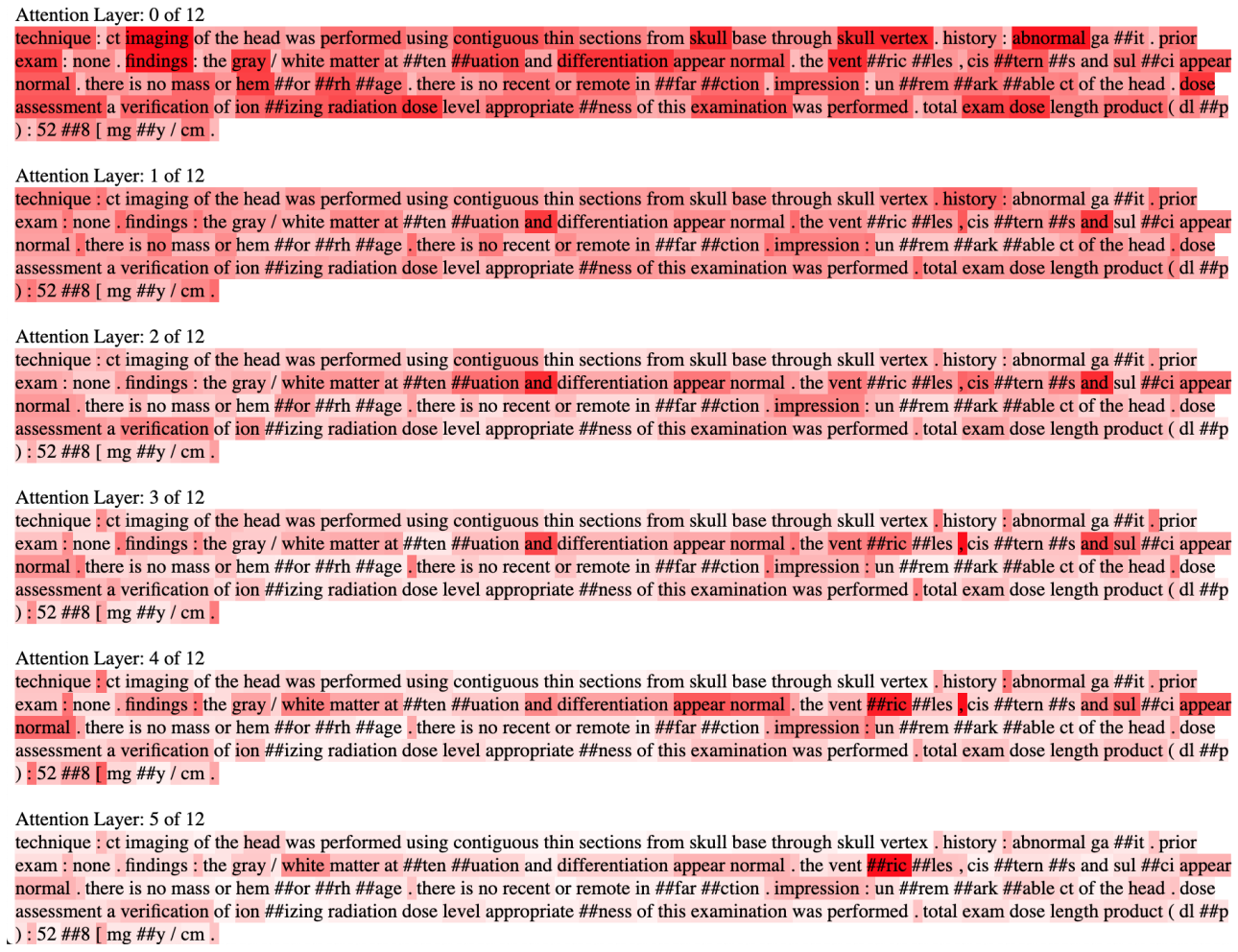

**Figure 4: Attention Visualization:** Heatmap of attention scores per token for the first 6 layers of the BERT model. Attention scores were accumulated for each token, and then normalized for the scores within one layer. The deeper the red shading the higher the attention score relative to the other attention scores within a single layer, indicating a higher emphasis on that token by the model.

The fine-tuning of the model parameters occurred over three epochs (the original BERT architecture paper recommends two – four[13]) on the training set of radiological reports for all three sets of parameters. In line with the methods outlined in the original BERT architecture paper,[13] all reports were converted to lowercase characters and each radiological report was tokenized using the wordpiece tokenizer, and appended with a start and stop sequence tag. Besides for these steps, no preprocessing was done to the reports. As suggested in the original BERT architecture paper,[13] the learning rate was $2e^{-5}$, the linear learning warmup proportion was 10%, and the training and test batch sizes were four reports each. The maximum radiological report sequence length was 300 tokens.

After each epoch of training, the model performance was evaluated on the evaluation set to ensure convergence. Model performance was measured by F1-score — defined as (two times precision times recall)/(precision plus recall), where precision refers to positive predictive value and recall refers to sensitivity — on a per-label basis and by a micro, macro, and sample-weighted average level. After three epochs were completed, the model was then evaluated on the same metrics as above on the test set, which included reports that the model had never yet seen. ROC curves and their respective AUC scores were generated for each model overall and for each label. 95% confidence intervals for each AUC score were calculated using a fast implementation of the DeLong algorithm[15]. The above process was repeated for both the general domain model, the BioBERT model, and the model using randomized parameters.

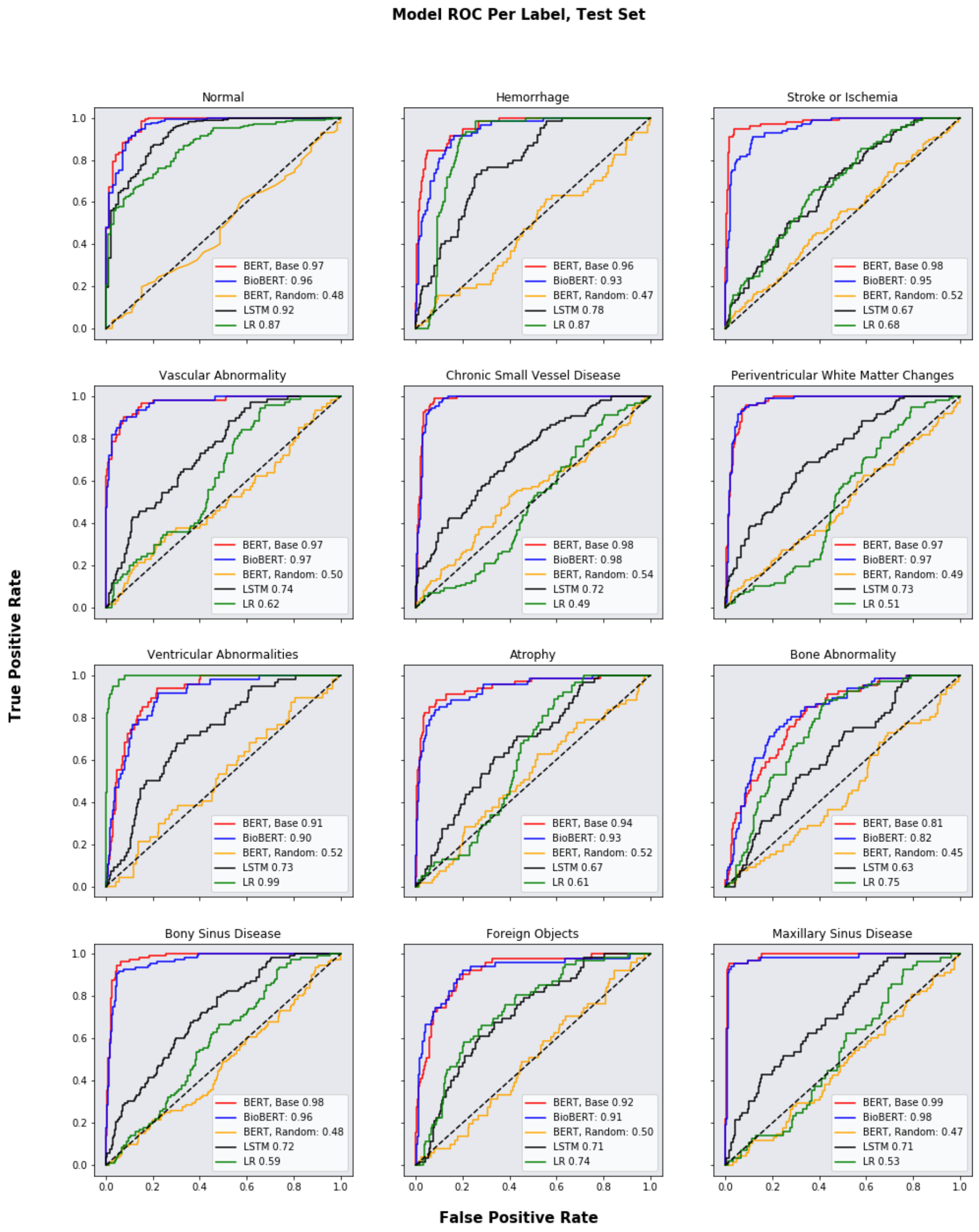

**Figure 5. Model Results:** Per-label receiver operator characteristic curves for the compared models: BERT pretrained on general domain corpora, BERT pretrained on general and medical domain corpora, BERT using randomized parameters, the LSTM model, and a logistic regression. The BERT models were superior for all labels, with the exception of Ventricular Abnormalities, where the logistic regression model outperforms the BERT models, and Bone Abnormalities, where the logistic regression model is comparable.

To benchmark performance of BERT, several comparison points were chosen. First, a simple multi-label multi-class logistic regression with a bag of words vectorization representation was run and evaluated on the same metrics. A logistic regression was chosen due to its widespread use in medical prediction tasks. Second, an LSTM using randomly initialized parameters was run on the same set of training, validation, and testing reports for 10 epochs (until convergence). The base LSTM was modified with a 13-dimensional output layer, and binary cross-entropy loss for multi-label multi-class classification. For all benchmark and index tests, a positive instance was specified as being a class probability of > 0.5.

## 3 Results

The labels included in the prediction task, based on having more than 250 positive instances, were as follows: normal (80% of all reports with positive instances), hemorrhage (13% of all reports with positive instances), stroke or infarction (24% of all reports with positive instances), vascular abnormality (17% of all reports with positive instances), chronic small vessel disease (29% of all reports with positive instances), periventricular white matter changes (27% of all reports with positive instances), ventricular abnormalities (13% of all reports with positive instances), brain atrophy (17% of all reports with positive instances), bone abnormality (17% of all reports with positive instances), bony sinus disease (26% of all reports with positive instances), foreign objects (15% of all reports with positive instances), maxillary sinus disease (15% of reports with positive instances), and carotid siphon calcification (13% of all reports with positive instances). These percentages are representative of the breakdown in label frequency across the test training and evaluation sets (see Table 1 for a complete breakdown).

The performance of the model in identifying the presence of each of the 13 labels was evaluated on the eval set after each epoch of training and was then evaluated on the held-out test set at the conclusion of training. The sample-weighted F1-score average of the various model types are as follows: 0.87 (general BERT), 0.87 (BioBERT), 0.39 (randomized BERT), 0.35 (LSTM), and 0.53 (LR). Results on a per-label and aggregate basis are summarized in Table 2.

In addition to the accuracy metrics summarized in Table 2, ROC curves and their corresponding AUC scores were generated to compare the models both on an aggregate and on a label-by-label basis (see Figure 5). The pretrained BERT models achieved an overall AUC score of 0.96 (95% CI: 0.95 - 0.97) with a range from 0.83 to 0.98 per label for the medical domain trained model, and overall AUC score of 0.97 (95% CI: 0.96 - 0.97) with a range of 0.81 to 0.99 per label for the general domain pretrained model. The randomized BERT model achieved an overall AUC score of 0.49 (95% CI: 0.47 – 0.50), with a range of 0.45 to 0.53 per label. The LSTM model achieved an overall AUC score of 0.79 (95% CI: 0.78 – 0.80) with a range of 0.63 to 0.91 per label. The logistic regression model had an overall ROC score of 0.64 (95% CI: 0.62 - 0.65) with a range of 0.49 to 0.99 per label.

**Table 1:** Label counts and frequencies broken down by the test, training, and eval datasets.

| | Test Set Label Counts | Test Set Label Frequency | Train Set Label Counts | Train Set Label Frequency | Eval Set Label Counts | Eval Set Label Frequency |
|---|---|---|---|---|---|---|
| Atrophy (Brain) | 62 | 15% | 215 | 19% | 67 | 17% |
| Bone Abnormality | 68 | 16% | 194 | 17% | 66 | 16% |
| Carotid Siphon Calcification | 53 | 13% | 156 | 13% | 47 | 12% |
| Chronic Small Vessel Disease | 128 | 31% | 334 | 29% | 110 | 27% |
| Foreign Objects | 62 | 15% | 180 | 16% | 51 | 13% |
| Hemorrhage | 60 | 15% | 132 | 11% | 57 | 14% |
| Maxillary Sinus Disease | 56 | 14% | 164 | 14% | 68 | 17% |
| Normal | 319 | 77% | 943 | 81% | 331 | 82% |
| Periventricular White Matter Changes | 119 | 29% | 314 | 27% | 99 | 25% |
| Sinus Disease (Bony, not Venous) | 108 | 26% | 296 | 25% | 112 | 28% |
| Stroke, Infarction, or Ischemia | 88 | 21% | 285 | 25% | 97 | 24% |
| Vascular Abnormality | 70 | 17% | 200 | 17% | 61 | 15% |
| Ventricular Abnormalities | 56 | 14% | 153 | 13% | 47 | 12% |

**Table 2:** F1, Precision, and Recall Performance (Test Set: N = 413 Reports)

| | F1 Score | | | | |
|---|---|---|---|---|---|
| Model Type | BERT | BioBERT | Random BERT | LSTM | Logistic Regression |
| Atrophy (Brain) | 0.77 | 0.78 | 0.90 | 0.94 | 0.47 |
| Bone Abnormality | 0.11 | 0.20 | 0.00 | 0.00 | 0.14 |
| Carotid Siphon Calcification | 0.91 | 0.88 | 0.39 | 0.00 | 0.46 |
| Chronic Small Vessel Disease | 0.90 | 0.91 | 0.00 | 0.00 | 0.77 |
| Foreign Objects | 0.50 | 0.71 | 0.00 | 0.02 | 0.38 |
| Hemorrhage | 0.56 | 0.52 | 0.00 | 0.02 | 0.44 |
| Maxillary Sinus Disease | 0.93 | 0.93 | 0.00 | 0.00 | 0.44 |
| Normal | 0.97 | 0.97 | 0.00 | 0.00 | 0.95 |
| Periventricular White Matter Changes | 0.84 | 0.83 | 0.00 | 0.00 | 0.69 |
| Sinus Disease (Bony, not Venous) | 0.90 | 0.87 | 0.00 | 0.00 | 0.59 |
| Stroke, Infarction, or Ischemia | 0.91 | 0.88 | 0.00 | 0.00 | 0.67 |
| Vascular Abnormality | 0.79 | 0.80 | 0.00 | 0.00 | 0.59 |
| Ventricular Abnormalities | 0.53 | 0.53 | 0.00 | 0.00 | 0.26 |
| **Sample-Weighted Average** | 0.87 | 0.87 | 0.39 | 0.35 | 0.53 |

## 4 Discussion

This study demonstrates the efficacy of transfer learning on general domain corpora for the classification of medical documents. We found that language models pretrained on general text achieve state of the art results on radiological text labelling, and, when compared with general domain pretraining, further pretraining with biomedical text provides no additional benefit to medical NLP tasks. These results may be of benefit in developing highly accurate clinical NLP models with limited access to labeled, deidentified medical corpora. We used the transformer-based BERT model with several sets of parameters, including a set pretrained on a general domain corpus, a set pretrained on a combined medical and general domain corpus, and a randomized set, for the subsequent classification of radiological documents.

Classifying the contents of a given medical note for large-scale research is a complex task from both a technical and an administrative standpoint. With no reliable metadata within the vast majority of EHRs today, researchers have to rely on either manual readers to review medical records or on billing codes that are poorly associated with ground truth labels[2–4]. Traditional NLP systems using rule-based classification may achieve impressive results on a given set of data, however significant human involvement is required to design these programs, and they generalize poorly to datasets outside those they were initially designed for. Published findings report that some rule-based systems achieve a positive predictive value of 0.45 – 0.75 on corpora other than those they were trained on[16]. Furthermore, more advanced NLP systems, including deep models capitalizing on transfer learning, have improved upon benchmark NLP results, however these models are not built in a domain-specific manner, and generalizability of models within medical machine learning is a known difficulty[9,13,18,19].

When using either set of pretrained parameters, BERT substantially outperformed the both the LSTM and the logistic regression on almost every measure. On the basis of F1-score, a combined measure of accuracy that factors in both precision and recall, the BERT models exceeded the baseline comparison models 0.87 and 0.87 respectively to 0.53 and 0.35 for the logistic regression and LSTM models. The low F1-score of the logistic regression model implies that both precision and recall are poor, meaning both comparison models had difficulties identifying true positives and true negatives. BERT's stronger performance than the LSTM and the logistic regression model is consistent across metrics, including ROC scores, and label types. There were specific instances in which the logistic regression model in particular outperformed BERT, namely for the ventricular abnormalities label, where the logistic regression model had the highest AUC, and bone abnormalities, where the logistic regression model was comparable in terms of AUC and F1-score. For bone abnormalities, all models had very poor F1-scores. This phenomenon is potentially due to the language of both ventricular and bone abnormalities being excessively heterogeneous, and the logistic regression model detecting select critical phrases that allow for the high precision and low recall we observed. The state-of-the-art results achieved by the pretrained BERT models, and the lack of efficacy when using randomized parameters, shows the ability of nonspecific text to generalize well to medical subdomain language tasks.

Of particular importance, the two varieties of pretrained parameters, general BERT and BioBERT, saw no significant differences for downstream tasks. On an overall-AUC basis, both parameters generated overlapping 95% confidence intervals, and the F1-scores on a per-label and an aggregate basis show few differences. As BioBERT was initialized with general domain pretraining followed by biomedical domain pretraining, this finding implies that the additional exposure of the model parameters to PubMed and PMC text had few benefits for our radiological classification task. There are several potential explanations for this finding. First, semantic research into the differences between biomedical literature and clinical notes found significant differences in the range of mentioned concepts, the frequency in which semantic concepts were invoked, and the generality of terms used[20]. Second, prior research on semantic modeling with unsupervised techniques suggests that beyond a certain point, the size of the corpus has little impact on the semantic representation gleaned[21]. These points suggest that the scientific text of PubMed and PMC contains little additional semantic information of value when applied in addition to the large general text that BERT is trained on. In order to enhance performance on a clinical predictive task, de-novo pretraining with a large corpus of medical text will likely be required.

While there is no standard medical domain-oriented NLP task with which to benchmark model performance, the existing literature suggests that transfer learning in conjunction with transformer-based models meets or exceeds the abilities of other NLP models in report labelling tasks. In one 2018 study comparing the efficiency of five machine learning algorithms, including Bayes point machine, logistic regression, random decision forest, neural network, and support vector machine algorithms, to classify CT and MR reports into one of four oncologic classes, none of the algorithms exceeded a F1-score of 0.80 in categorizing the full report[22]. In comparison, BERT shows significantly better F1-scores at 0.87 for both sets of pretrained parameters in a 13-class problem. Furthermore, BERT achieved the higher F1-score on a dataset that consisted of roughly 20% of the size (1,977 reports vs 9,418 reports) of the corpus used in the 2018 study[22]. These findings further bolster the argument that general domain pretraining is a powerful technique to achieve state of the art results in a data-limited environment.

As with all scientific studies, there are important limitations to our work. One limitation is the generalizability of our trained model to other datasets. The exact classifiers derived from training on the report corpus in this work may not generalize to external data sets within neurology due to variance in reporting style and would certainly struggle to perform on non-neurological CT scans, even if acquired at the same site[19]. Unique EHR templates for reporting, as well as language conventions specific to a hospital system, can prevent reasonable performance on datasets acquired in separate locations. Additionally, because of the very specific nature of the vocabulary associated with neuropathology, this specific model, would likely fail to perform when considering pathology associated, for example, with spine or abdominal findings. Given the versatile nature of language models, BERT would likely achieve similarly excellent results on other medical tasks requiring an equivalent level of linguistic understanding, provided the dataset contains enough accurately labeled samples.

Second, medical datasets continue to be a substantial bottleneck in applying NLP techniques on a broader basis to radiological reports, as well as medicine as a whole. Labeled data is extremely time-consuming to generate and is an almost universal concern of machine learning and deep learning researchers. Domain transfer techniques, such as the unsupervised pre-training techniques adopted in the BERT model, serve to bolster performance on datasets of limited size by endowing language models with a broad understanding of language before approaching any specific task, however these methods cannot entirely replace representative data. Especially in medicine, where heterogenous sets of etiologies and descriptors can reference the same symptoms, and where the language in a note or report can serve the dual purpose of liability protection as well the description of a condition, NLP models can struggle to accurately approximate the true data distribution[23]. These considerations manifested themselves in our work by the exclusion of the vast majority of the labels due to heavy class imbalance (see Figure 2), as well as the inability of any of the model variants to perform well when considering ventricular and bone abnormalities. Solutions such as crowd-sourcing labels or the development of autonomous silver-standard labelling methods show promise but have yet to be developed into capable solutions to this problem.

Lastly, sequence modeling architectures and language models all contain limitations when applied to medical text. The first of which is that all models currently contain a limit to how much continuous text can be passed to them. The original implementation of BERT is limited to a maximum of 512 continuous tokens, which is substantial enough to encompass the vast majority of the relatively brief radiology reports in our corpus. If longer sets of text are used, such as an indolent hospital course in a discharge summary or a complicated radiology report with multiple incidental findings, the length of the note could far exceed the number of tokens BERT can handle.

## 5 Conclusions

In conclusion, general text used for transfer learning is a viable technique to generate state-of-the-art results within medical NLP tasks on radiological corpora, outperforming other deep models such as LSTMs. The efficacy of pretraining and transformer-based models could serve to facilitate the creation of groundbreaking

NLP models in the uniquely challenging data environment of medical text. Further improvements on clinical subdomain tasks will likely require novel parameters pretrained first on text specific to the task at hand.

## REFERENCES

[1] Bochicchio, M. et al. 2016. A Big Data Analytics Framework for Supporting Multidimensional Mining over Big Healthcare Data. *2016 15th IEEE International Conference on Machine Learning and Applications (ICMLA)* (Dec. 2016), 508–513.

[2] Carrell, D.S. et al. 2014. Using natural language processing to improve efficiency of manual chart abstraction in research: the case of breast cancer recurrence. *American journal of epidemiology*. 179, 6 (Mar. 2014), 749–758.

[3] Chen, P.-H. et al. 2018. Integrating Natural Language Processing and Machine Learning Algorithms to Categorize Oncologic Response in Radiology Reports. *Journal of digital imaging*. 31, 2 (Apr. 2018), 178–184.

[4] Devlin, J. et al. 2018. BERT: Pre-training of Deep Bidirectional Transformers for Language Understanding. *arXiv [cs.CL]*.

[5] Floyd, J.S. et al. 2012. Use of administrative data to estimate the incidence of statin-related rhabdomyolysis. *JAMA: the journal of the American Medical Association*. 307, 15 (Apr. 2012), 1580–1582.

[6] Garla, V. et al. 2011. The Yale cTAKES extensions for document classification: architecture and application. *Journal of the American Medical Informatics Association: JAMIA*. 18, 5 (Sep. 2011), 614–620.

[7] Hassanpour, S. et al. 2017. Performance of a Machine Learning Classifier of Knee MRI Reports in Two Large Academic Radiology Practices: A Tool to Estimate Diagnostic Yield. *AJR. American journal of roentgenology*. 208, 4 (Apr. 2017), 750–753.

[8] Jorge, A. et al. 2019. Identifying lupus patients in electronic health records: Development and validation of machine learning algorithms and application of rule-based algorithms. *Seminars in arthritis and rheumatism*. (Jan. 2019). DOI:https://doi.org/10.1016/j.semarthrit.2019.01.002.

[9] Lee, J. et al. 2019. BioBERT: a pre-trained biomedical language representation model for biomedical text mining. *arXiv [cs.CL]*.

[10] Liang, H. et al. 2019. Evaluation and accurate diagnoses of pediatric diseases using artificial intelligence. *Nature medicine*. (Feb. 2019). DOI:https://doi.org/10.1038/s41591-018-0335-9.

[11] Miller, J.M. et al. 2010. The natural language of the surgeon's clinical note in outcomes assessment: a qualitative analysis of the medical record. *The American Journal of Surgery*.

[12] Misky, G.J. et al. 2018. Hospital Readmission From the Perspective of Medicaid and Uninsured Patients. *Journal for healthcare quality: official publication of the National Association for Healthcare Quality*. 40, 1 (2018), 44–50.

[13] Murff, H.J. et al. 2011. Automated identification of postoperative complications within an electronic medical record using natural language processing. *JAMA: the journal of the American Medical Association*. 306, 8 (Aug. 2011), 848–855.

[14] Pakhomov, S.V.S. et al. 2016. Corpus domain effects on distributional semantic modeling of medical terms. *Bioinformatics* . 32, 23 (Dec. 2016), 3635–3644.

[15] Peng, Y. et al. 2019. Transfer Learning in Biomedical Natural Language Processing: An Evaluation of BERT and ELMo on Ten Benchmarking Datasets. *arXiv [cs.CL]*.

[16] Percha, B. et al. 2012. Automatic classification of mammography reports by BI-RADS breast tissue composition class. *Journal of the American Medical Informatics Association: JAMIA*. 19, 5 (Sep. 2012), 913–916.

[17] Solti1, I. et al. 2009. Automated classification of radiology reports for acute lung injury: Comparison of keyword and machine learning based natural language processing approaches. *2009 IEEE International Conference on Bioinformatics and Biomedicine Workshop* (Nov. 2009), 314–319.

[18] Sun, X. and Xu, W. 2014. Fast Implementation of DeLong's Algorithm for Comparing the Areas Under Correlated Receiver Operating Characteristic Curves. *IEEE Signal Processing Letters*. 21, 11 (Nov. 2014), 1389–1393.

[19] Turchin, A. et al. 2009. Comparison of information content of structured and narrative text data sources on the example of medication intensification. *Journal of the American Medical Informatics Association: JAMIA*. 16, 3 (May 2009), 362–370.

[20] Turchin, A. et al. 2006. Using regular expressions to abstract blood pressure and treatment intensification information from the text of physician notes. *Journal of the American Medical Informatics Association: JAMIA*. 13, 6 (Nov. 2006), 691–695.

[21] Wu, S. and Liu, H. 2011. Semantic characteristics of NLP-extracted concepts in clinical notes vs. biomedical literature. *AMIA ... Annual Symposium proceedings / AMIA Symposium. AMIA Symposium*. 2011, (Oct. 2011), 1550–1558.

[22] Zech, J. et al. 2018. Natural Language–based Machine Learning Models for the Annotation of Clinical Radiology Reports. *Radiology*. 287, 2 (May 2018), 570–580.

[23] Zech, J.R. et al. 2018. Variable generalization performance of a deep learning model to detect pneumonia in chest radiographs: A cross-sectional study. *PLoS medicine*. 15, 11 (Nov. 2018), e1002683.